\documentclass{article}

\usepackage{arxiv}

\usepackage[utf8]{inputenc} 
\usepackage[T1]{fontenc}    
\usepackage{hyperref}       
\usepackage{url}            
\usepackage{booktabs}       
\usepackage{amsfonts}       
\usepackage{nicefrac}       
\usepackage{microtype}      
\usepackage{lipsum}
\usepackage{color}
\usepackage{graphicx}
\usepackage{subcaption}
\usepackage{amsmath} 
\usepackage{amsxtra}
\usepackage{amssymb}
\usepackage{amsthm}
\usepackage{algorithm, algpseudocode}
\usepackage[table,xcdraw]{xcolor}
\usepackage{array,ragged2e,booktabs,colortbl} 
\usepackage{multicol,multirow}
\usepackage{hhline}
\usepackage{latexsym}
\usepackage{siunitx}

\newcommand{\func}{\textsc}
\newcommand{\ternary}[3]{#1 \text{ ? } #2 : #3}
\newcommand{\concat}[2]{#1 \text{ \& } #2}
\let\oldReturn\Return
\renewcommand{\Return}{\State\oldReturn}
\newcommand{\positenv}[2]{Posit$\langle #1, #2 \rangle$}

\title{Template-Based Posit Multiplication for Training and Inferring in Neural Networks}

\author{
  Raúl Murillo Montero \\
  Department of Computer Architecture and Automation\\
  Complutense University of Madrid\\
  Madrid, 28040 \\
  \texttt{ramuri01@ucm.es} \\
   \And
  Alberto A. Del Barrio \\
  Department of Computer Architecture and Automation\\
  Complutense University of Madrid\\
  Madrid, 28040 \\
  \texttt{abarriog@ucm.es} \\
   \AND
  Guillermo Botella \\
  Department of Computer Architecture and Automation\\
  Complutense University of Madrid\\
  Madrid, 28040 \\
  \texttt{gbotella@ucm.es} \\
}

\begin{document}
\maketitle

\begin{abstract}
The posit number system is arguably the most promising and discussed topic in Arithmetic nowadays. The recent breakthroughs claimed by the format proposed by John L. Gustafson have put posits in the spotlight. In this work, we first describe an algorithm for multiplying two posit numbers, even when the number of exponent bits is zero. This configuration, scarcely tackled in literature, is particularly interesting because it allows the deployment of a fast sigmoid function. The proposed multiplication algorithm is then integrated as a template into the well-known FloPoCo framework. Synthesis results are shown to compare with the floating point multiplication offered by FloPoCo as well. Second, the performance of posits is studied in the scenario of Neural Networks in both training and inference stages. To the best of our knowledge, this is the first time that training is done with posit format, achieving promising results for a binary classification problem even with reduced posit configurations. In the inference stage, 8-bit posits are as good as floating point when dealing with the MNIST dataset, but lose some accuracy with CIFAR-10.
\end{abstract}


\section{Introduction}
\label{sec:introduction}

Multiple floating-point representations have been used in computers over the years, although the IEEE Standard for Floating-Point Arithmetic (IEEE 754) \cite{IEEE754_1985} is the most common implementation that modern computing systems have adopted. Since it was established in 1985, the standard has only been revisited in 2008 (IEEE 754-2008) \cite{IEEE_Std}, but the main characteristics of the original remain to keep compatibility with existing implementations and it is not adopted by all computer systems. However, multiple shortcomings have been identified in the IEEE 754 standard, which are listed below \cite{gustafson2017end}:

\begin{itemize}
	\item Different computers using the same IEEE floating-point format are not required produce the same results. When a computation does not fit into the chosen number representation, the number will be rounded. Even in the last revision of the standard they introduce the \emph{round-to-nearest, ties away from zero} rounding scheme and provide recommendations for computations reproducibility, hardware designers are not coerced to implement them. Therefore, identical computations can lead to multiple results across different computing platforms \cite{kahan1998java}.
	\item Multiple bit patterns are used for handling exceptions such as the \emph{Not a Number} (NaN) value, which indicates that a value is not representable or undefined -- for example dividing by zero results in a NaN. The problem is that the amount of bit patterns that represent NaN may be more than necessary, making hardware design more complex and decreasing the available number of exactly representable values.
	\item IEEE 754 makes use of overflow -- accepting $\infty$ or $-\infty$ as a substitute for large-magnitude finite numbers -- and underflow -- accepting 0 as a substitute for small-magnitude nonzero numbers. Thus, major problems can be produced, as the above mentioned.
	\item Rounding is performed on individual operands of every calculation, so associativity and  distributivity properties are not always held in floating-point representation. The last revision of the standard tries to solve this issue including the Multiply Accumulation (FMA) operation. However, again this may not be supported by all computer systems.
\end{itemize}

The above listed shortcomings led to the idea of developing a new number system that can serve as a replacement for the now ubiquitous IEEE 754 arithmetic. At the beginning of 2017, John L. Gustafson introduced the \emph{posit} number representation system, a Type III unum format \cite{gustafson2017end,Jaiswal_2018_adder} that has no underflow, overflow or wasted NaN values. Gustafson claims that posits are not only a suitable replace for the current IEEE Standard for Floating-Point Arithmetic, but also provide more accurate answers with an equal or smaller number of bits and simpler hardware \cite{gustafson2017beating}. As it is illustrated in \cite{gustafson2017beating,gustafson2017posit,gustafson2016radical} there are important benefits when using posits, as a better dynamic range, accuracy, closure and consistency between machines than with conventional floating point. However, posits are still in development and there is still some controversy about their improvement \cite{deDinechin2019posits,ughen2019HAL}.

\begin{figure}
    \centering
    \includegraphics[height=1.2in]{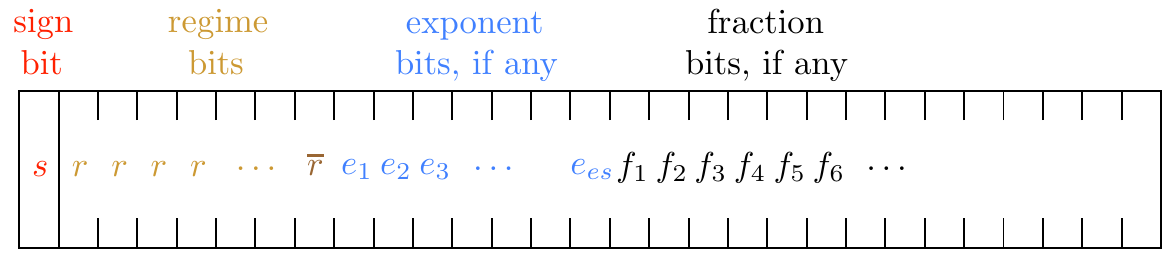}
    \caption{Layout of an $\langle n,es \rangle$ posit number.}
    \label{fig:posit_format}
\end{figure}

The numerical value of a posit number $X$, whose bits are distributed as shown in Figure \ref{fig:posit_format}, is given by Equation \ref{eq:posit_decimal_value}:

\begin{equation}
    \label{eq:posit_decimal_value}
    X = (-1)^{s}*(2^{2^{es}})^{k}*2^{e}*1.f~,
\end{equation}{}

where $k$ is the \emph{regime} value, $e$ is the unsigned exponent (if $es$ > 0), and $f$ is the mantissa of the number without the implicit one. In terms of format layout, the main differences with floating point are the existence of the regime field and together with the unsigned and unbiased exponent, if there exists such exponent field. The regime is a sequence of bits with the same value (\emph{r}) finished with the negation of such value
($\overline{r}$). Provided that $X=x_{n-1}x_{n-2}...x_{1}x_{0}$, this regime can be expressed as Equation \ref{eq:posit_regime} shows.

\begin{equation}
    \label{eq:posit_regime}
    k = - x_{n-2} + \sum_{i=n-2}^{x_{i}\neq~x_{n-2}}(-1)^{1 - x_{i}}~.
\end{equation}

In other words, the regime basically counts the number of occurrences of the bit labelled as $r$ in Figure \ref{fig:posit_format}. If $r=1$ then the regime is the number of 1's minus 1, while if $r=0$, the regime is the negative value of the number of 0's. For instance, if the regime is 4-bits wide, the value \texttt{1110} would be interpreted as $k=2$, while the value \texttt{0001} would be $k=-3$. Hence, detecting the leading 1's and 0's is critical for performing this step \cite{Kim_2018,Kim_2019,tavares2019design}. However, the main difficulty to detect the regime, and consequently unpack the posit, is that its length varies dynamically. In numbers close to zero, there will be many fraction bits and few regime bits and, in large numbers there will be many regime bits and fewer fraction bits, but there is no fixed amount of bits. Finally, it must be noted that the scaling factor defined by the regime ($k$) and the amount of exponent bits ($es$) is typically named $useed$. 

This variability of configurations that the posit numbers provide is an excellent opportunity to research in the design of efficient units able to implement operations among posits. The recommended formats for obtaining similar accuracy results with respect to floating point are:  $\langle 8,0 \rangle$ (w.r.t \emph{minifloat} \cite{minifloat}),  $\langle 16,1 \rangle$ (w.r.t. half precision) and  $\langle 32,2 \rangle$ (w.r.t. single precision). Nevertheless, the aforementioned variability of posit configurations opens a gate to look for functional units providing the best trade-offs \cite{Barrio2019,barrio2017slack}. It is therefore desirable to provide a generic architecture and a generic flow to implement posit functional units. For this purpose, in this paper we leverage the FloPoCo framework \cite{de_Dinechin_2011} capabilities. FloPoCo (Floating-Point Cores, but not only) is an open-source C++ framework for the generation of arithmetic datapaths which provides a command-line interface that inputs operator specifications and outputs synthesizable VHDL specially suited for Field-Programmable Gate Arrays (FPGAs). In this paper we have defined a generic posit multiplication algorithm and integrated it as a template within the FloPoCo framework. This way, it is possible to generate synthesizable multipliers with any posit configuration, including $es=0$. The code is publicly available at github \footnote{\url{https://github.com/RaulMurillo/Posit-Multiplier_FloPoCo}}. Moreover, we have evaluated the performance of this algorithm and other posit operators in one of the scenarios where they can have a greater impact because of their accuracy with reduced formats: the Neural Networks (NNs). To the best of our knowledge, this is the first time that posits are evaluated in the training stage, providing promising results. In the inference phase, short posit formats match floating point for the MNIST dataset and lose some accuracy with CIFAR-10.

The rest of the paper is organized as follows: Section \ref{sec:related work} describes the state of the art regarding posit multipliers and other functional units; Section \ref{sec:posit multiplier} presents our algorithm for multiplying in the posit format as well as its integration with the FloPoCo framework; Section \ref{sec:posits and nns} presents a study on posits and NNs; Section \ref{sec:experiments} provides synthesis and simulation results to validate the approach and finally Section \ref{sec:conclusions} draws our conclusions and future lines of work.

\section{Related Work}
\label{sec:related work}

Since the posit number system was introduced, the interest on a hardware implementation for this format has increased rapidly. Despite the short life time of posits, several hardware implementations have been proposed since 2017.

The posit arithmetic unit proposed in \cite{Jaiswal_2018,Jaiswal_2018_adder}, includes floating-point to posit conversion, posit to floating-point conversion, addition/subtraction and multiplication. The work in \cite{Thesis:2018} improves the decimal accuracy of the latters and defines a posit vectorized unit. Although these works seem totally parametrized, there is no support for the case of zero exponent bits. This is important because in the Deep Learning scenario, the configuration $\langle 8,0 \rangle$ provides an extremely fast approach for the sigmoid function \cite{gustafson2017posit}. 

Another general design for posit arithmetic unit that includes adders and multipliers is presented in \cite{Chaurasiya_2018}. In contrast to the implementations shown in \cite{Jaiswal_2018, Thesis:2018}, the posit decoder proposed in this architecture uses only a leading zero detector for decoding the regime, while \cite{Jaiswal_2018,Jaiswal_2018_adder} were using a leading one detector too. The work in \cite{jaiswal2019pacogen} improves the prior ones by employing just one leading one detector and provides a tool for generating different units with different posit formats. Authors in \cite{podobas2018hardware} present a C++ template compliant with Intel OpenCL SDK. Nonetheless, the configuration with $es=0$ is not included either in any of the aforementioned articles. J. Johnson \cite{Johnson2018} employs posit addition to complement the logarithmic multiplication when performing inference in Convolutional Neural Networks (CNNs), studying the  $\langle 8,0 \rangle$ configuration too, but just in simulation. Other studies tackling the inference stage of CNNs are performed in \cite{carmichael2019date,carmichael2019performance,Langroudi2018PositNNTP}. In all these works, the training stage is performed in floating point and the weights converted to posit format, while the inference stage is performed in posit format.

Finally, for the sake of completeness it must be noted that since the posit standard \cite{posit_standard} includes fused operations such as the fused dot product, and due to the importance of this operation for matrix calculus, some research and development for this kind of implementations has been done. Different matrix-multiply units for posits are presented in \cite{Chen_2018, Thesis:2018, carmichael2019date, carmichael2019performance}. They make use of the \emph{quire} register \cite{posit_standard} to accumulate the partial additions that are involved in the dot product, so the result is rounded only after the whole computation. Nevertheless, the fused operations are out of the scope of this work, as the functional unit that will be described is a posit multiplier.

In this paper we propose an algorithm to perform the multiplication of two posit numbers and integrate it into a well-known framework as FloPoCo \cite{de_Dinechin_2011}, providing support for generating synthesizable multipliers for any posit configuration, including the case $es=0$. Furthermore, we present a study in both the training and inference stages of NNs. As it there have been mentioned, there have been several studies for the inference stage, but not on the training phase.

\section{The Posit Multiplier}
\label{sec:posit multiplier}

At hardware level, posits were designed to be easy to compute, i.e., to have a circuitry similar to the existing floating point. The main encoding difference between float and posit formats is the fact that the second one includes a run-time varying scaling component. This leads to a format that has no fixed fields at run-time, which is a hardware design challenge. Below we present a fully functional posit multiplier operator.

Analogously to performing computations with IEEE floats, it is necessary to unpack/decode the operands fields before carrying out any computation. Therefore, we first present the posit decoding process in Algorithm \ref{alg:1}. It must be noted that the decoder can also be used in other arithmetic modules, e.g. a posit adder. The explanation of such algorithm is as follows:
\begin{itemize}
	\item Sign and special cases are detected checking the Most Significant Bit (MSB) and ORing the remaining bits, respectively (lines 2--5).
	\item Since posit arithmetic uses 2's complement for representing negative numbers, dealing with the absolute value simplifies the data extraction process. Therefore, the 2's complement of the inputs are obtained, only if it is necessary, by XORing the input with the replicated sign bit  and adding the sign to the Least Significant Bit (LSB) (line 6).
	\item The $twos[N-2]$ bit aids to determine the regime value. In order to use only a leading zero detector \cite{Chaurasiya_2018}, we invert the bits of $twos$ if the regime consists on a sequence of ones (line 8). Then we count the sequence of \texttt{0} bits terminating in a \texttt{1} bit using a leading zero detector module (line 9).
	\item For extracting the exponent and the fraction bits, the regime is shifted out from $twos$, so the exponent is aligned to the left (line 10).
	\item The first $es$ bits of the shifted string (if $es=0$ this instruction is omitted) correspond to the exponent bits (line 11), while the remaining bits correspond to the fraction (line 12). It must be noted that here the hidden bit is appended as the MSB. 
	\item The regime depends on the sequence of identical bits that constitute this field. The regime value is $zc-1$ when the bits are \texttt{1} (positive regime) or $-zc$ when it consists on a sequence of \texttt{0} bits (negative regime). Note that an extra \texttt{0} is added to maintain sign bit of the operation (line 13), as $zc$ is a positive value.
\end{itemize}

\begin{algorithm}[hbt]
	\caption{Posit data extraction}\label{alg:1}
	\begin{algorithmic}[1]
		\Procedure{Decode}{$in$}
		\State $nzero \gets \bigvee in[N-2:0]$ \Comment{Reduction OR}
		\State $sign \gets in[N-1]$	\Comment{Extract sign}
		\State $z \gets \neg (sign \vee nzero)$
		\State $inf \gets sign \wedge \neg (nzero)$
		\State $twos \gets (\{N-1 \{sign\} \} \oplus in[N-2:0]) + sign$	\Comment{Input 2's complement}
		\State $rc \gets twos[N-2]$	\Comment{Regime check}
		\State $inv \gets \{N-1\{rc\}\} \oplus twos$
		\State $zc \gets \func{LZD}(inv)$	\Comment{Count leading zeros of regime}
		\State $tmp \gets twos[N-4:0] \ll (zc-1)$	\Comment{Shift out the regime}
		\State $exp \gets tmp[N-4:N-es-3]$	\Comment{Extract exponent}
		\State $frac \gets \concat{nzero}{tmp[N-es-4:0]}$	\Comment{Extract fraction}
		\State $reg \gets \ternary{rc}{\concat{\texttt{`0'}}{zc-1}}{-(\concat{\texttt{`0'}}{zc})}$	\Comment{Select regime}
		\Return $sign, reg, exp, frac, z, inf$
		\EndProcedure
	\end{algorithmic}
\end{algorithm}

The process of posit multiplication is almost the same as for floating-point multiplication, i.e. the scaling factors are added and the fractions are multiplied and rounded. There are few differences when multiplying posits due to the regime field. The pseudocode for posit multiplication is shown in Algorithm \ref{alg:2} and the explanation of the flow is as follows:

\begin{algorithm}[htb!]
	\caption{Posit Multiplier Algorithm}\label{alg:2}
	\begin{algorithmic}[1]
		\Procedure{PositMult}{$in_A, in_B$}
		\State $sign_A, reg_A, exp_A, frac_A, z_A, inf_A \gets \func{Decode}(in_A)$
		\State $sign_B, reg_B, exp_B, frac_B, z_B, inf_B \gets \func{Decode}(in_B)$
		\State $sign \gets sign_A \oplus sign_B$	\Comment{Sign computation}
		\State $z \gets z_A \vee z_B$		\Comment{Special cases computation}
		\State $inf \gets inf_A \vee inf_B$
		\State $sf_A \gets \concat{reg_A}{exp_A}$ \Comment{Gather scale factors}
		\State $sf_B \gets \concat{reg_B}{exp_B}$
		\State $frac_{mult} \gets frac_A \times frac_B$	\Comment{Fractions multiplication}
		\State $ovf_m \gets frac_{mult}[MSB]$		\Comment{Adjust for overflow}
		\State $norm_{frac} \gets \ternary{ovf_m}{\concat{\texttt{`0'}}{frac_{mult}}}{\concat{frac_{mult}}{\texttt{`0'}}}$	\Comment{Normalize fraction}
		\State $sf_{mult} \gets (\concat{sf_A[MSB]}{sf_A}) + (\concat{sf_B[MSB]}{sf_B}) + ovf_m$	\Comment{Add scaling factors}
		\State $sf_{sign} \gets sf_{mult}[MSB]$	\Comment{Get regime's sign}
		\State $nzero \gets \bigvee frac_{mult}$
		\State $exp \gets sf_{mult}[es-1:0]$	\Comment{Unpack scaling factors}
		\State $reg_{tmp} \gets sf_{mult}[MSB-2:es]$
		\State $reg \gets \ternary{sf_{sign}}{- reg_{tmp}}{reg_{tmp}}$	\Comment{Get regime's absolute value}
		\State $ovf_{reg} \gets reg[MSB]$	\Comment{Check for regime overflow}
		\State $reg_f \gets \ternary{ovf_{reg}}{\concat{\texttt{`0'}}{\{\lceil\log_2(N)\rceil\{\texttt{`1'}\} \}}}{reg}$	
		\State $ovf_{regf} \gets \bigwedge reg_f[MSB-2:0]$
		\State $exp_f \gets \ternary{(ovf_{reg} \vee ovf_{regf} \vee \neg nzero)}{\{es\{\texttt{`0'}\}\}}{exp}$
		\State $tmp1 \gets \concat{nzero}{\concat{\texttt{`0'}}{\concat{exp_f}{\concat{norm_{frac}[MSB-3:0]}{\{N-1\{\texttt{`0'}\} \}}}}}$	\Comment{Packing}
		\State $tmp2 \gets \concat{\texttt{`0'}}{\concat{nzero}{\concat{exp_f}{\concat{norm_{frac}[MSB-3:0]}{\{N-1\{\texttt{`0'}\} \}}}}}$
		\State $shift_{neg} \gets \ternary{ovf_{regf}}{reg_f - 2}{reg_f - 1}$
		\State $shift_{pos} \gets \ternary{ovf_{regf}}{reg_f - 1}{reg_f}$
		\State $tmp \gets \ternary{sf_{sign}}{tmp2 \gg shift_{neg}}{tmp1 \gg shift_{pos}}$	\Comment{Final answer with extra bits}
		\State $LSB, G, R \gets tmp[MSB-(N-1):MSB-(N+1)]$ \Comment{Unbiased rounding}
		\State $S \gets \bigvee tmp[MSB-(N+2):0]$
		\State $round \gets \ternary{(ovf_{reg} \vee ovf_{regf})}{\texttt{`0'}}{G \wedge (LSB \vee R \vee S)}$
		\State $result_{tmp} \gets {\concat{\texttt{`0'}}{(tmp[MSB:MSB-(N-1)] + round)}}$
		\State $result \gets \ternary{inf}{infinity}{\ternary{z}{zero}{\ternary{sign}{- result_{tmp}}{result_{tmp}}}}$
		\Return $result$
		\EndProcedure
	\end{algorithmic}
\end{algorithm}

\begin{itemize}
	\item When the two operands are decoded (lines 2--3), the sign and special cases are handled easily (lines 4--6).
	\item The Scaling Factor (SF) of each operand consists of the regime and the exponent values, one after the other (lines 7--8). This is due to how posit decimal values are computed using regime and exponent.
	\item The resulting fraction field is the outcome after multiplying the two operands fractions as if they were integer values (line 9). Recall that multiplying two $n-bit$ integers results in an integer of $2n$ bits. In addition, the decoder module returns fractions with the hidden bit as MSB, so the first two bits of the fractions multiplication do not strictly belong to the fraction field of the result, since they correspond to the multiplication of the hidden bits plus the possible carry bit due to fraction overflow. Therefore, the MSB of the result aids to detect any overflow when multiplying the fractions(line 10).
	\item If a fraction overflow occurs, the resulting fraction has to be normalized shifting one bit to the right. In order to avoid losing any bit for rounding, instead of shifting, we just append a \texttt{0} bit as MSB, or as LSB if there is no overflow (line 11).
	\item The resulting scaling factor is obtained by adding both operand scales, plus the possible fraction overflow. The result of adding two bit strings of same size may overflow, and in this case that carry bit indicates the sign of resulting regime, so it is necessary to replicate the MSB of both scaling factors before adding them (lines 12--13).
	\item Exponent and regime are extracted from the scaling factors addition. The obtained regime may be negative, but it is more suitable to handle absolute values (lines 15--17). A similar action is performed in the decoding stage to simplify the following steps.
	\item Adding two high-magnitude regimes may result in overflow, so in that case the regime is truncated to the maximum possible value and the exponent is set to 0 (lines 18--21).
	\item Once the resulting fields have been computed and adjusted, they have to be packed in the correct order. To construct the regime correctly, the packed fields have to be right-shifted as a signed integer according to the sign and value of the regime. It is important to avoid losing any fraction bit to round correctly, so an amount of \texttt{0} bits has to be appended on the right (lines 22--26).
	\item Posits, same as IEEE 754 floats, follow a \emph{round-to-nearest-even} scheme. To perform a correct unbiased rounding, the LSB, Guard (G), Round (R) and Sticky (S) bits are needed \cite{Koren:1993} (lines 27--29). The rounded result is finally adjusted according to the sign and exceptions.
\end{itemize}

\subsection{Integration with FloPoCo}
\label{subsec:integration flopoco}

In this subsection we can briefly comment how to integrate the aforementioned algorithms within the FloPoCo framework \cite{de_Dinechin_2011}. FloPoCo follows an object-oriented class hierarchy, where all operators inherit from a baseline \verb|Operator| virtual class. Thus, by extending such class and incorporating Algorithms \ref{alg:1} and \ref{alg:2} it is possible to create a posit multiplier with $n$ and $es$ as input parameters.

\begin{figure}[h]
	\centering
	\begin{subfigure}{0.46\textwidth}
		\centering
		\includegraphics[width=\textwidth]{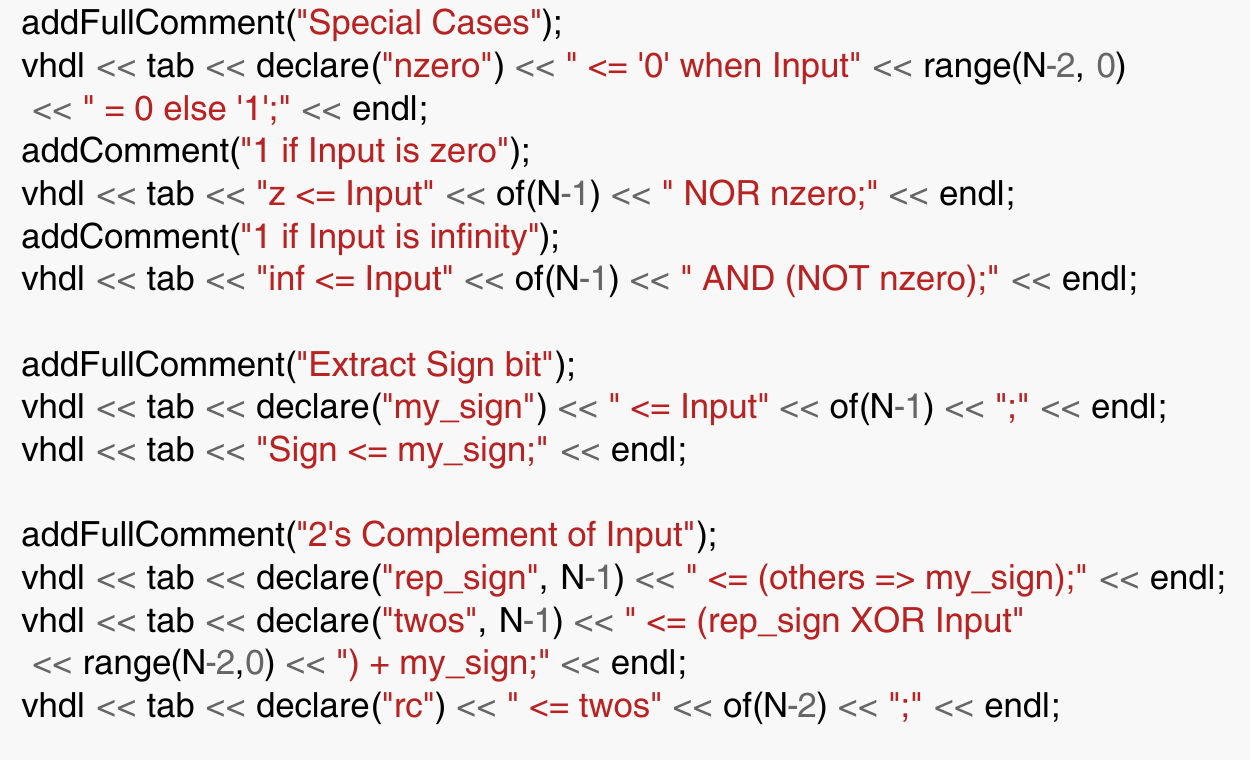}
		\caption{Source code in \texttt{PositMult.cpp} file.}
	\end{subfigure}
	\quad
	\begin{subfigure}{0.46\textwidth}
		\centering
		\includegraphics[width=\textwidth]{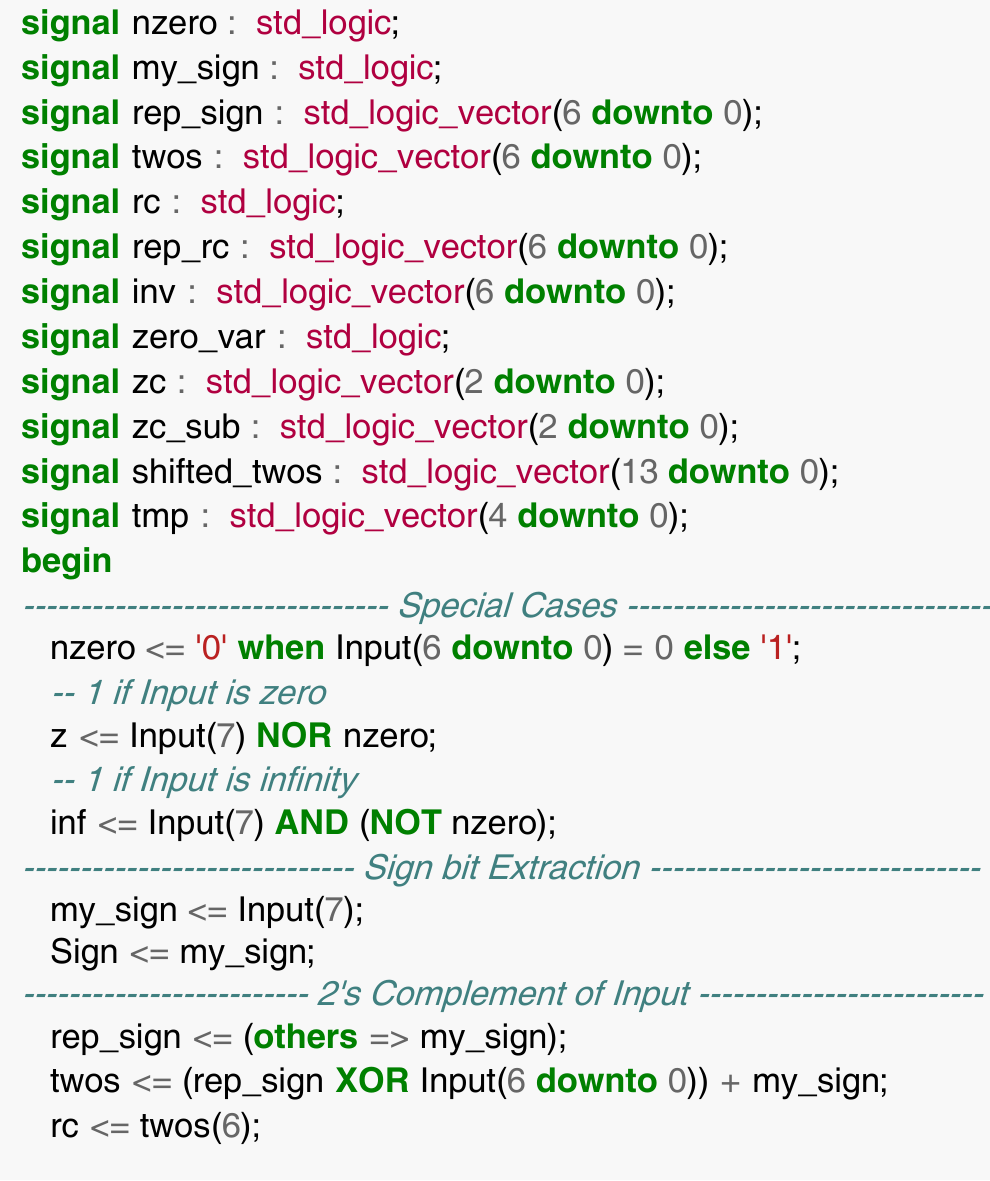}
		\caption{Generated VHDL code.}
	\end{subfigure}
	\caption{Generation of synthesizable VHDL from C++ code with FloPoCo.}
	\label{fig:FloPoCo}
\end{figure}

Then, using the command \verb|flopoco <options> <operator specification list>|, FloPoCo will generate a single synthesizable VHDL file \cite{de_Dinechin_2011}. Figure \ref{fig:FloPoCo} illustrates this process for the command \verb|flopoco PositMult N=8 es=1|,  with which we obtain the VHDL code for a $\langle 8,1 \rangle$ multiplier, and changing the values on \verb|N| and \verb|es| we can obtain a new multiplier for any other posit configuration.

It is important to mention that, in contrast with the works presented in \cite{Jaiswal_2018, Jaiswal_2018_adder, Chaurasiya_2018} which only provide implementations with a non-zero value for $es$, we designed a generic template that can be used to automatically generate multipliers for any posit configuration, not only those with $es > 0$. What is more, it is possible to generate combinational and sequential and even FPGA-customized versions of the multiplier by just changing the options when invoking FloPoCo. This will be shown in Section \ref{sec:experiments}.

\section{Case Study: Posits and Neural Networks}
\label{sec:posits and nns}

The recent surge of interest in Artificial Intelligence, and in particular in Deep Learning (DL), together with the limitations this sector currently has in terms of power consumption and memory resources make us wonder if posits can be helpful in this field. As described in this paper, the posit number system has many interesting properties, such as lack of underflow or overflow or the fast approximation of sigmoid function that some configurations of posits can do ($es=0$). These, along with the so-called \textit{tapered precision} \cite{Langroudi2018PositNNTP}, suggest that posits may be suitable for performing DL tasks.

\begin{figure}[h]
	\centering
	\begin{subfigure}{0.475\textwidth}
		\centering
		\includegraphics[width=\textwidth]{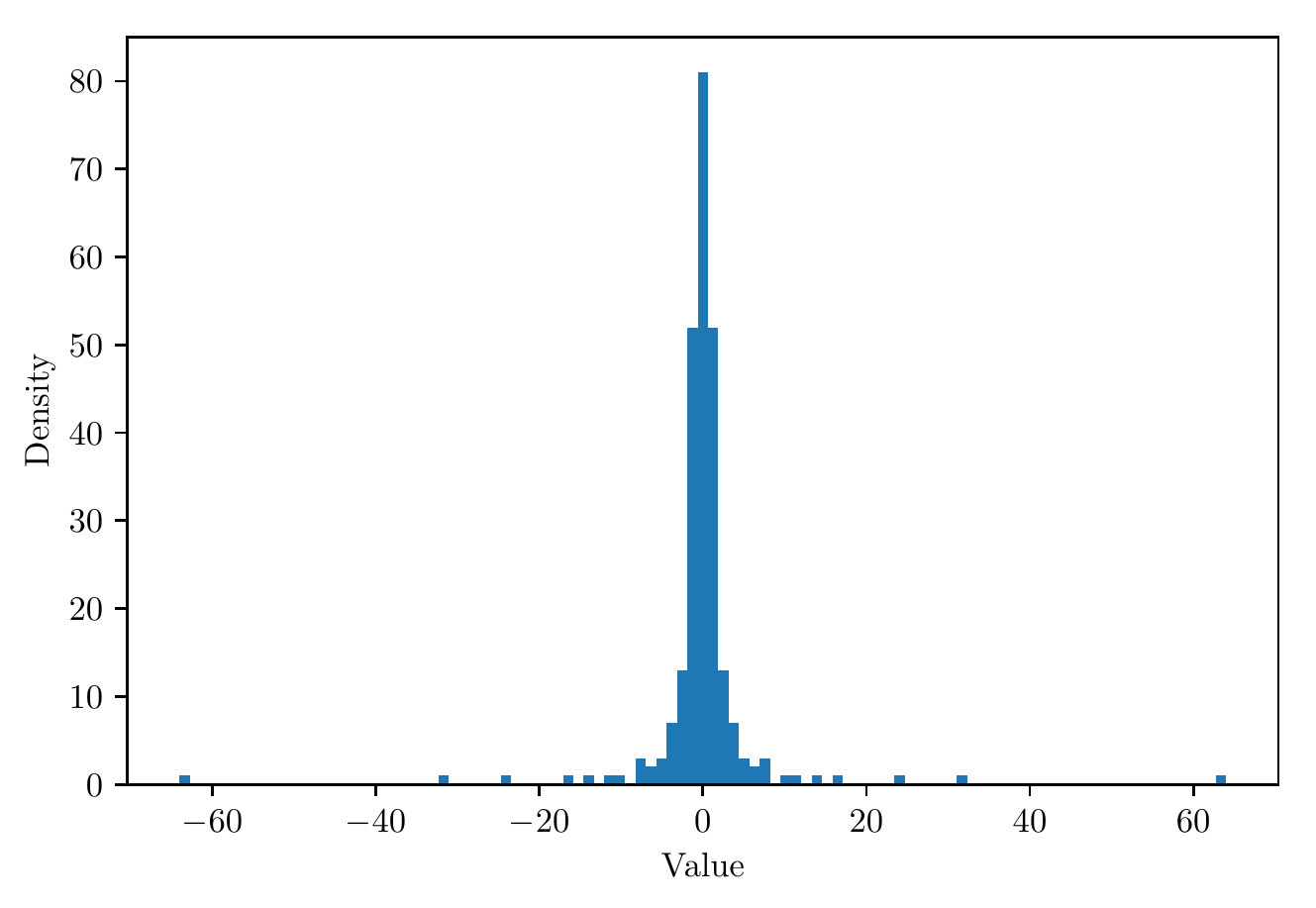}
		\caption {Distribution of $\langle 8,0 \rangle$ values.}
	\end{subfigure}
	\hfill
	\begin{subfigure}{0.475\textwidth}
		\centering
		\includegraphics[width=\textwidth]{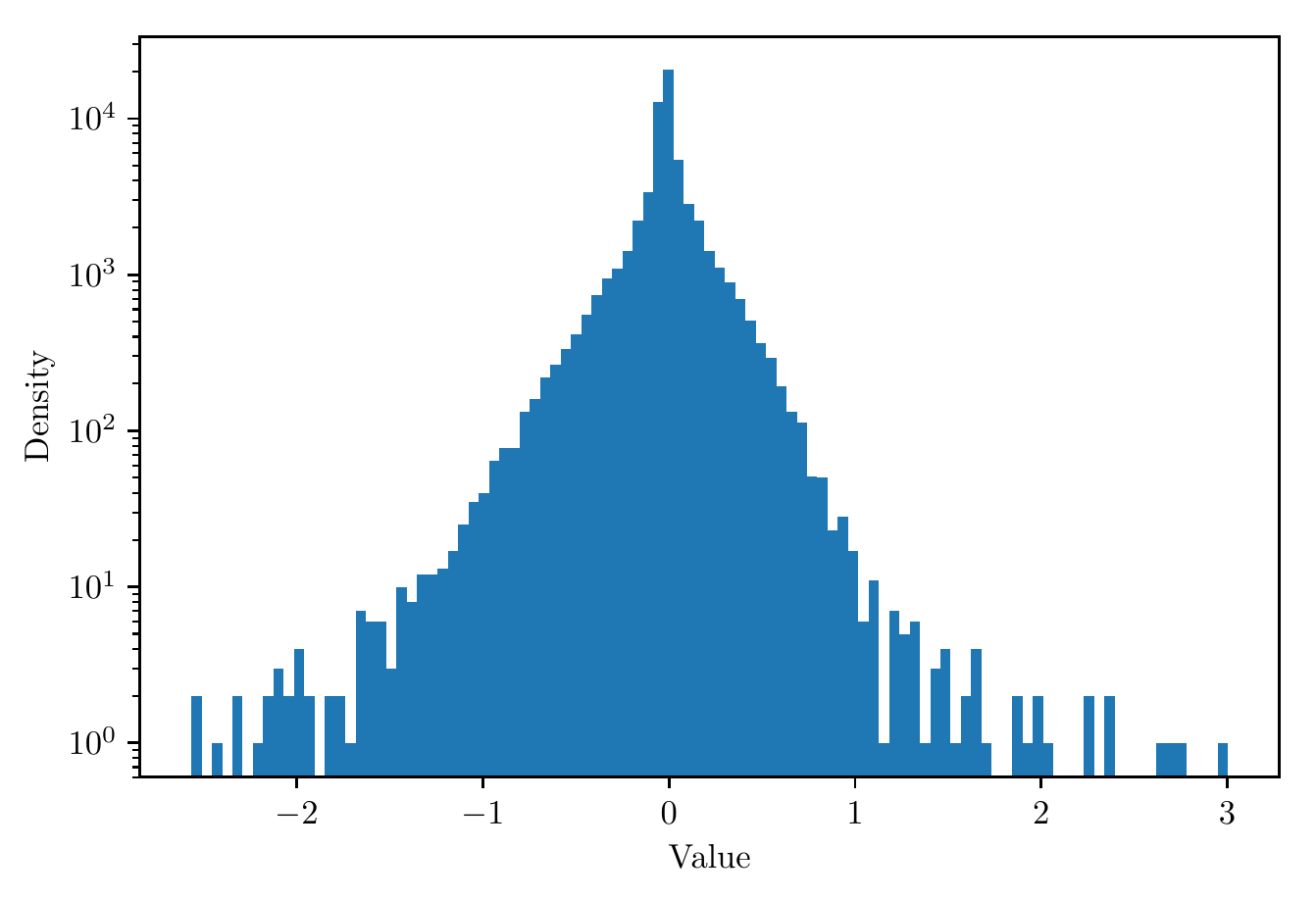}
		\caption {LeNet-5 weight distribution for CIFAR-10.}
	\end{subfigure}
	\caption{Distributions of posit values and NN weights.}
	\label{fig:tapered}
\end{figure}

In a format with tapered precision the values mass around 0 and sparse to higher or lower numbers in less frequency, so representation of small values is more accurate than using other formats. When we use a number system with tapered precision, such as posits, the values follow a normal distribution centered in 0. That is the same distribution that DNN weight parameters usually follow, but even more grouped around 0. Figure \ref{fig:tapered} illustrates this concept, which suggests that using posits for DNN may provide more accurate results.

\subsection{Training}
\label{subsec:training}

The NN training with the posit format has been done using the posit-arithmetic library PySigmoid \cite{pysigmoid}. The choice of this particular package is due to the fact that it allows working with specific posit configurations, not only the ``common'' \positenv{8}{0}, \positenv{16}{1} and \positenv{32}{2}, and that it has a function that simulates the hardware operation for fast sigmoid, which approximates the original function when posits have $es=0$, in particular when using the \positenv{8}{0} configuration. According to \cite{gustafson2017posit}, this fast sigmoid can be achieved by flipping the first bit of the posit and shifting it right two places, so given an $n$-bits posit $X$, this behavior can be modeled by Equation \ref{eq:fast_sigmoid},  describes how the fast sigmoid is implemented for an $n$-bit input.

\begin{equation}
    \label{eq:fast_sigmoid}
    \sigma_{fast}(X,n) = X \oplus 2^{n-1}>> 2~.
\end{equation}

\begin{figure}[t]
    \centering
    \includegraphics[width=0.6\textwidth]{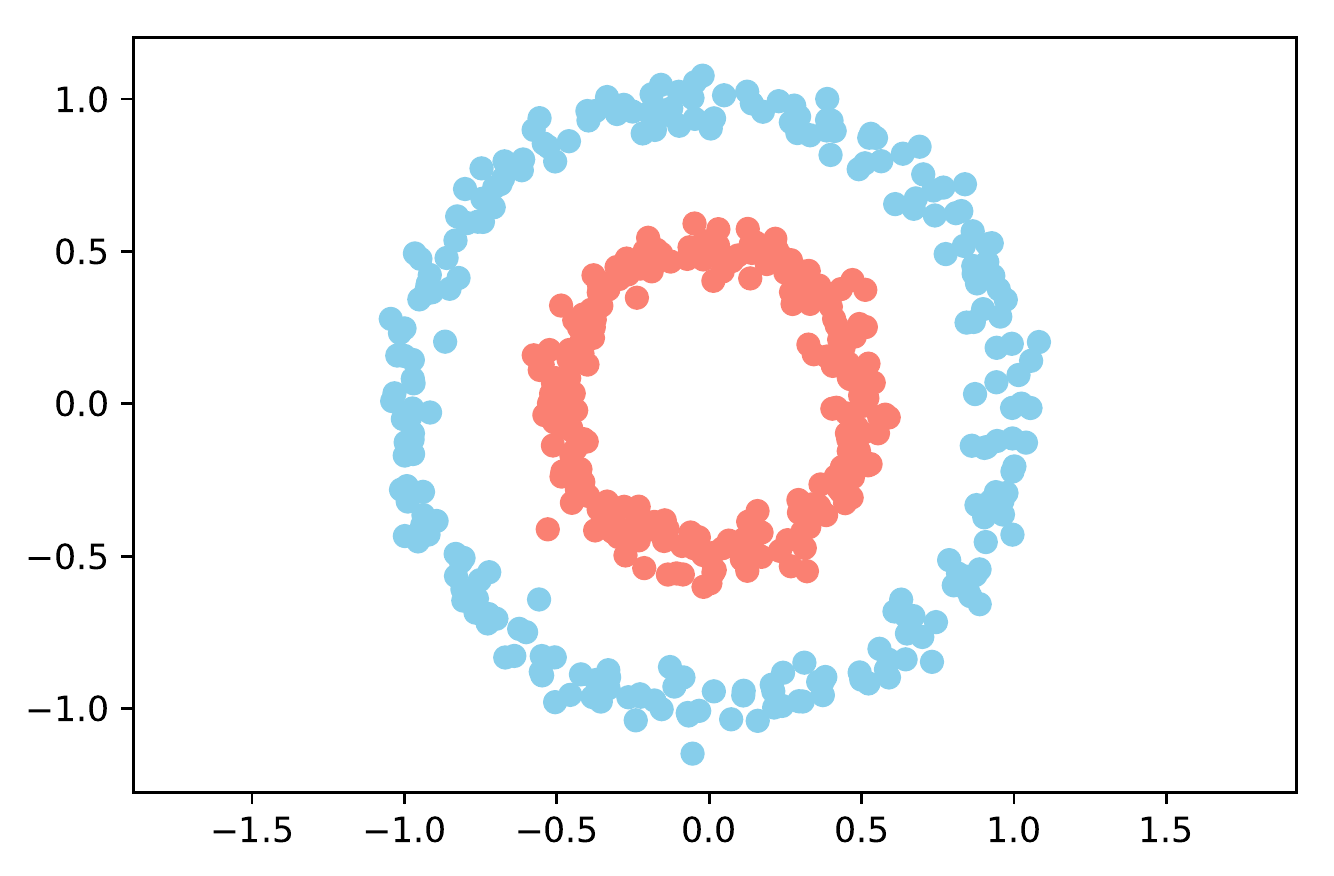}
    \caption{Classification problem for posit training.}
    \label{fig:dataset}
\end{figure}

To measure how well posits perform at deep learning tasks, the binary classification problem depicted in Figure \ref{fig:dataset} has been studied in detail. The samples consist only of two features and classes are obviously separated by a non-linear boundary.

The case study consists in training a NN architecture with two hidden layers of 4 and 8 neurons, respectively, to solve the aforementioned binary classification problem and check if posits may be suitable for training. Although there are multiple libraries and frameworks for Machine Learning (ML) that accelerate and simplify these kind of tasks, our test requires that all the internal computations involving parameters of the network are done in the posit format. Hence, the only option is to implement the NN from scratch, casting the input into posit type and replacing all the internal operands by the ones from PySigmoid library \cite{pysigmoid}. In this way the fused dot product with the quire accumulator can be employed too. In order to perform operations with the quire, the original library was modified and can be found at github \footnote{\url{https://github.com/RaulMurillo/PySigmoid/tree/master/PySigmoid}}. The original version of this library allowed underflow for certain values, which was fixed in the modified version we present in this paper.

\subsection{Inference}
\label{subsec:inference}

As some research papers show \cite{Chen2014, Gupta2015, Courbariaux2014}, it is difficult to apply lower numerical precision to the training of NNs, especially when using less than 16 bits. However, many research papers have shown that it is possible to apply low-precision computing to the inference stage of NNs after training with exact arithmetic \cite{rodriguez2018lower, Jacob2017, Hubara2016}. These results lead to the idea of using a reduced posit format such as \positenv{8}{0} for carrying out the DL inference. Performing low-precision inference can be extremely helpful in embedded systems and applications that make use of DL techniques such as Autonomous Driving \cite{Cococcioni2018}.

There are some characteristics of the posit format that can be an advantage when using this format in NNs:
\begin{itemize}
	\item The comparison of posits uses the same hardware as for comparing integers, which is much faster than floats comparison. Thus, the typical pooling layers of CNNs could be easily implemented.
	\item If using a format with $es=0$, e.g. \positenv{8}{0}, the sigmoid function can be approximately calculated as described by Equation \ref{eq:fast_sigmoid}.
	\item Input values for NNs are usually normalized between [-1,1]. Because of the aforementioned tapered precision, the addition of two posit numbers is pretty accurate near 0.
\end{itemize}

\section{Experiments}
\label{sec:experiments}

In this section the experimental results are presented. First, the synthesis results of template-based multiplier are discussed and second, the performance of posits operators is evaluated on NNs in both training and inference stages.

\subsection{Synthesis Results}
\label{subsec:synthesis}

Prior to synthesizing the posit multiplier described in Section \ref{sec:posit multiplier}, the verification has been done as follows: first the golden solution has been obtained with the help of the Mathematica environment \cite{mathematica2019}. Afterwards, the VHDL the test bench is run using Xilinx Vivado Design Suite \cite{vivado} and the outputs compared with those from Mathematica, producing no mismatch. Different posit configurations have been successfully tested. In particular, complete tests have been done for $\langle 8,0 \rangle$, $\langle 8,1 \rangle$ and $\langle 8,2 \rangle$, and also, but less exhaustive, for $\langle 16,1 \rangle$ and $\langle 32,2 \rangle$.

Several posit multipliers have been synthesized using Synopsys Design Compiler with a 65 \si{\nano\meter} target-library and without placing any timing constraint. The delay, area, power and energy of different posit multipliers have been measured. Besides synthesizing multipliers for different posit formats, the capabilities of FloPoCo have been leveraged to generate these units with different styles, namely: pipelined, combinational and combinational without employing hard multipliers nor DSP blocks. Also, several floating-point multipliers have been generated using FloPoCo as well. In this case, the notatio $\langle exp,mant \rangle$ indicates the amount of bits for the exponent and mantissa bits, respectively. Given that there is also a sign bit, the three explored FloPoCo floating point configurations would correspond with minifloat ($\langle 4,3 \rangle$), IEEE Half Precision ($\langle 5,10 \rangle$) and IEEE Single Precision ($\langle 8,23 \rangle$). Nevertheless, it must be noted that FloPoCo floating point does not handle exceptional cases (infinity, NaN) and subnormal numbers. Table \ref{tab:synthesis} presents the synthesis results. In case of the pipelined designs, the number of stages is indicated between parenthesis next to the delay value.

\begin{table}[htb]
	\centering
	\scalebox{0.85}{
	\begin{tabular}{ll|lllll|lll|}
		
		\hhline{*2~|*5{-}|*3{-}|}
		& \multicolumn{1}{l|}{}                    & \multicolumn{5}{c|}{\positenv{n}{es} configurations} &    \multicolumn{3}{c|}{FP $\langle exp,mant \rangle$ configurations}                                                                                                                             \\ \hhline{*2~|*5{-}|*3{-}|}
		
		&                                          & $\langle 8,0 \rangle$               & $\langle 8,1 \rangle$       & $\langle 8,2 \rangle$       & $\langle 16,1 \rangle$      & $\langle 32,2 \rangle$ & $\langle 4,3 \rangle$ & $\langle 5,10 \rangle$ & $\langle 8,23 \rangle$               \\ \hline
		\multicolumn{1}{|l|}{}                                               & \cellcolor[HTML]{EFEFEF}Pipelined            & \cellcolor[HTML]{EFEFEF}$0.8$ (8)     & \cellcolor[HTML]{EFEFEF}$0.79$ (8)   & \cellcolor[HTML]{EFEFEF}$0.78$ (7)    & \cellcolor[HTML]{EFEFEF}$1.06$ (10)  & \cellcolor[HTML]{EFEFEF}$2.3$ (15) 
		& \cellcolor[HTML]{EFEFEF}$0.63$ (2)  & \cellcolor[HTML]{EFEFEF}$1.07$ (3) 
		& \cellcolor[HTML]{EFEFEF}$1.58$ (2)   
		\\
		\multicolumn{1}{|l|}{}                                               & Combinational                                 & $3.59$                             & $3.52$                             & $3.17$                             & $6.2$                              & $10.34$                              &
		$1.18$ & $2.62$ & $4.97$\\
		\multicolumn{1}{|l|}{\multirow{-3}{*}{Delay (\si{\nano\second})}}                  & \cellcolor[HTML]{EFEFEF}Combinational,  No hm & \cellcolor[HTML]{EFEFEF}$3.36$     & \cellcolor[HTML]{EFEFEF}$3.23$     & \cellcolor[HTML]{EFEFEF}$3.18$     & \cellcolor[HTML]{EFEFEF}$6.2$      & \cellcolor[HTML]{EFEFEF}$9.6$      & \cellcolor[HTML]{EFEFEF}$1.18$      &
		\cellcolor[HTML]{EFEFEF}$2.62$      & \cellcolor[HTML]{EFEFEF}$4.39$      
		\\ \hline
		\multicolumn{1}{|l|}{}                                               & Pipelined                                    & $2799$                            & $2745$                            & $2481$                            & $6898$                            & $24299$ & $950$ & $3800$ & $7757$
		\\
		\multicolumn{1}{|l|}{}                                               & \cellcolor[HTML]{EFEFEF}Combinational         & \cellcolor[HTML]{EFEFEF}$1488$    & \cellcolor[HTML]{EFEFEF}$1483$    & \cellcolor[HTML]{EFEFEF}$1415$    & \cellcolor[HTML]{EFEFEF}$3865$    & \cellcolor[HTML]{EFEFEF}$15459$    &
		\cellcolor[HTML]{EFEFEF}$700$    & \cellcolor[HTML]{EFEFEF}$2883$     &
		\cellcolor[HTML]{EFEFEF}$6684$         
		\\
		\multicolumn{1}{|l|}{\multirow{-3}{*}{Area (\si{\micro\squared\metre})}} & Combinational,  No hm                         & $1271$                            & $1152$                            & $1048$                            & $3865$                            & $21894$                             &
		$700$ & $2883$ & $11640$
		\\ \hline
		\multicolumn{1}{|l|}{}                                               & \cellcolor[HTML]{EFEFEF}Pipelined            & \cellcolor[HTML]{EFEFEF}397      & \cellcolor[HTML]{EFEFEF}$384.3$    & \cellcolor[HTML]{EFEFEF}$313.7$    & \cellcolor[HTML]{EFEFEF}$862.1$    & \cellcolor[HTML]{EFEFEF}$2269.2$     &
		\cellcolor[HTML]{EFEFEF}$158.6$    & \cellcolor[HTML]{EFEFEF}$815.3$ &
		\cellcolor[HTML]{EFEFEF}$4362.8$
		\\
		\multicolumn{1}{|l|}{}                                               & Combinational                                 & $631.3$                            & $562.1$                            & $428.4$                            & $2609.6$                           & $12693.6$                            &
		$121.4$ & $987.9$ & $4983.4$
		\\
		\multicolumn{1}{|l|}{\multirow{-3}{*}{Power (\si{\micro\watt})}}                   & \cellcolor[HTML]{EFEFEF}Combinational,  No hm & \cellcolor[HTML]{EFEFEF}$612.4$    & \cellcolor[HTML]{EFEFEF}$503.9$    & \cellcolor[HTML]{EFEFEF}$424$      & \cellcolor[HTML]{EFEFEF}$2609.6$   & \cellcolor[HTML]{EFEFEF}$13053.3$    & \cellcolor[HTML]{EFEFEF}$121.4$   &
		\cellcolor[HTML]{EFEFEF}$987.9$    & \cellcolor[HTML]{EFEFEF}$4956.5$  
		\\ \hline
		\multicolumn{1}{|l|}{}                                               & Pipelined                                    & $2.54$                           & $2.43$                         & $1.71$                         & $9.14$                         & $78.29$ & $0.20$ & $2.62$ & $13.79$                           
		\\
		\multicolumn{1}{|l|}{}                                               & \cellcolor[HTML]{EFEFEF}Combinational         & \cellcolor[HTML]{EFEFEF}$2.27$ & \cellcolor[HTML]{EFEFEF}$1.98$ & \cellcolor[HTML]{EFEFEF}$1.36$ & \cellcolor[HTML]{EFEFEF}$16.18$ & \cellcolor[HTML]{EFEFEF}$131.25$ &
		\cellcolor[HTML]{EFEFEF}$0.14$ & \cellcolor[HTML]{EFEFEF}$2.59$ &
		\cellcolor[HTML]{EFEFEF}$24.77$ 
		\\
		\multicolumn{1}{|l|}{\multirow{-3}{*}{Energy (\si{\pico\joule})}}                  & Combinational,  No hm                         & $2.06$                         & $1.63$                         & $1.35$                          & $16.18$                         & $125.31$ &
		$0.14$ & $2.59$ & $21.76$
		\\ \hline
	\end{tabular}
	}
	\caption{Posit multipliers synthesis results.} 
	\label{tab:synthesis}
\end{table}

A first conclusion that can be extracted is the fact that posit pipelined designs are not optimized in terms of pipeline depth, as they have many stages in comparison with equivalent floating point units. For example, 8-bit multipliers require at least 7 stages, which is a lot for this kind of components. Second, the FloPoCo units are more efficient but, as it has been mentioned before, they are not complete while our multiplier is.

In order to compare results with a state-of-the-art multiplier \cite{Chaurasiya_2018}, we have used Xilinx Vivado for synthesizing our proposed multiplier on a ZedBoard Zynq-7000 SoC, the same target as in \cite{Chaurasiya_2018}. This comparison in terms of LUTs and DSPs is shown in Table \ref{tab:FPGA}. In the case of the pipelined design, it must be also considered that extra resources are necessary. For example, for the case of 32-bits, 65 LUTRAMs, 910 FFs and one BUFG are required. 

\begin{table}[htb]
	\centering
	\begin{tabular}{|l|l|l|l|l|}
		\hline
		& \multicolumn{2}{c|}{\positenv{16}{1}} & \multicolumn{2}{c|}{\positenv{32}{2}}  \\ \hhline{|~*2{|--}}
		\multirow{-2}{*}{Datapath} & Slice LUT    & DSP   & Slice LUT    & DSP   \\ \hline
		\textbf{\cite{Chaurasiya_2018}}        & 218 & 1 & 572 & 4 \\ \hline
		Pipelined                 & 321          & 1          & 891          & 2          \\ \hline
		Combinational              & 266          & 1          & 927          & 2          \\ \hline
		Combinational, No hm       & 266          & 1          & 1640         & 0          \\ \hline
	\end{tabular}
	\caption{Comparison of posit multipliers synthesis area results.} \label{tab:FPGA}
\end{table}

As can be observed in Table \ref{tab:FPGA}, the results are not as good as the ones presented in \cite{Chaurasiya_2018}. Nonetheless, it must be observed that for 32-bits our implementations employ less DSPs. Furthermore, it must be reminded that our flow also supports the case of $es=0$. 

\subsection{Neural Networks Training}
\label{subsec:neural networks training}

As it has been described in Section \ref{subsec:training}, different \positenv{n}{es} configurations has been studied for the binary classification problem. Moreover, the posit configurations employing $es=0$ have utilized the fast sigmoid function too, while those with $es>0$ or floating point formats have made use of the regular sigmoid. The weights and biases are randomly initialized and the Mean Square Error (MSE) has been used as loss function to compare the outcomes of the network.

\begin{figure}[t]
    \centering
    \includegraphics[width=0.8\textwidth]{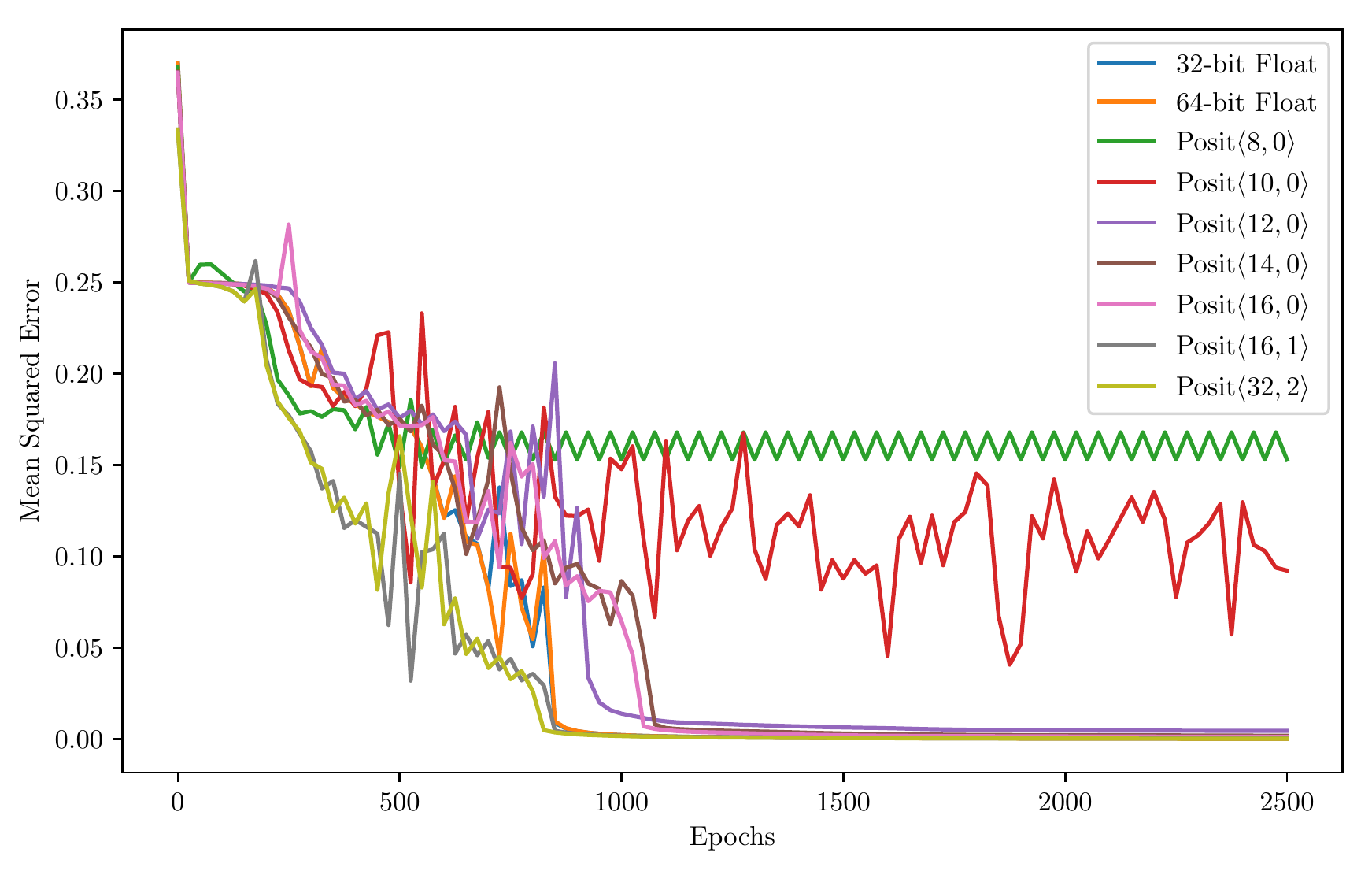}
    \caption{Loss function along the NN training.}
    \label{fig:nn}
\end{figure}

An amount of 2500 epochs have been set to compare the losses of the different formats throughout the whole training. In this manner it is possible to compare whether the network converges or not and also how fast. Table \ref{tab:nn} and Figure \ref{fig:nn} depict the obtained results.

\begin{table}[htb]
	\centering
	\begin{tabular}{l|l|l|l|l|l|l|}
		\hhline{*1~|*6{-}|}
		& \multicolumn{6}{c|}{\cellcolor[HTML]{EFEFEF}Epochs} \\ \hline
		\rowcolor[HTML]{EFEFEF} 
		\multicolumn{1}{|l|}{\cellcolor[HTML]{EFEFEF}Configuration} & 0    & 250    & 500    & 750    & 1000    & 1250    \\ \hline
		\multicolumn{1}{|l|}{32-bit Float}                          & 0.3701 & 0.2346 & 0.1726 & 0.0839 & 0.0023 & 0.0010 \\ \hline
		\multicolumn{1}{|l|}{64-bit Float}                          & 0.3701 & 0.2346 & 0.1727 & 0.1124 & 0.0023 & 0.0010 \\ \hline
		\multicolumn{1}{|l|}{\positenv{8}{0}}     & 0.3681 & 0.1882 & 0.1491 & 0.1530 & 0.1530 & 0.1530 \\ \hline
		\multicolumn{1}{|l|}{\positenv{10}{0}}    & 0.3653 & 0.2129 & 0.1359 & 0.0938 & 0.1478 & 0.1264 \\ \hline
		\multicolumn{1}{|l|}{\positenv{12}{0}}    & 0.3650 & 0.2467 & 0.1758 & 0.1684 & 0.0140 & 0.0081 \\ \hline
		\multicolumn{1}{|l|}{\positenv{16}{0}}    & 0.3648 & 0.2817 & 0.1716 & 0.1622 & 0.0645 & 0.0035 \\ \hline
		\multicolumn{1}{|l|}{\positenv{16}{1}}    & 0.3337 & 0.1772 & 0.1453 & 0.0440 & 0.0019 & 0.0011 \\ \hline
		\multicolumn{1}{|l|}{\positenv{32}{2}}    & 0.3337 & 0.1758 & 0.1658 & 0.0328 & 0.0017 & 0.0009 \\ \hline
	\end{tabular}
\caption{Loss function during the NN training.} \label{tab:nn}
\end{table}

As can be seen, there is almost no difference in using single or double precision floats. Both \positenv{32}{2} and \positenv{16}{1} present the same behavior as floats, even with less MSE during the fists epochs. Posit configurations with 16, 14 and 12 bits takes some extra epochs to converge, and only those configurations with 10 and 8 bits present an irregular convergence. In these cases, the lack of underflow is undermining the posits convergence. In fact, this type of behavior has already appeared in a Newton-Raphson study presented at \cite{bachelor2018}.

Therefore, it can be concluded that posits have converged as well as floats, even with some short formats employing the fast sigmoid approach. Although the proposed NN is a reduced example, these facts point in a good direction to train more complex CNNs.

\subsection{Neural Networks Inference}
\label{subsec:neural networks inference}

The performance of \positenv{8}{0} format is evaluated on two datasets: MNIST and CIFAR-10 run on the LeNet-5 architecture \cite{Lecun1998}. In this case, the networks are firstly trained with floating point arithmetic and then the weights are converted to posit format prior to the inference stage. The networks have been trained using Keras \cite{chollet2015keras} and TensorFlow \cite{tensorflow2015-whitepaper} frameworks. The posit computations during inference have been simulated with the help of a NumPy library version which includes a posit data type \cite{numpy}. In this case, computations are much faster than using the PySigmoid library, but there is not a fast sigmoid implementation, so the simpler ReLU module is used as activation function instead. The obtained results are shown in Table \ref{tab:CNN}.

\begin{table}[htb]
	\centering
	\begin{tabular}{l|r|r|r|r|r|r|}
		\hhline{~|*6{-}|}
		& \multicolumn{2}{c|}{\cellcolor[HTML]{EFEFEF}32-bit Float}                                               & \multicolumn{2}{c|}{\cellcolor[HTML]{EFEFEF}\positenv{8}{0}}                         & \multicolumn{2}{c|}{\cellcolor[HTML]{EFEFEF}\begin{tabular}[c]{@{}c@{}}\positenv{8}{0}\\ (only addition)\end{tabular}} \\ \hline
		\rowcolor[HTML]{EFEFEF} 
		\multicolumn{1}{|l|}{Dataset} & Top-1 & Top-5 & Top-1 & Top-5 & Top-1 & Top-5 \\ \hline
		\multicolumn{1}{|l|}{MNIST}              & $99.22\%$                                            & $100.00\%$                                           & $99.32\%$                                            & $99.94\%$                                            & $99.40\%$                                            & $100\%$                                              \\ \hline
		\multicolumn{1}{|l|}{CIFAR-10}           & $68.04\%$                                            & $96.47\%$                                            & $56.11\%$                                            & $92.42\%$                                            & $58.92\%$                                            & $95.62\%$                                            \\ \hline
	\end{tabular}
\caption{Performance on CNNs inference.} \label{tab:CNN}
\end{table}

As can be observed, posits get slightly higher Top-1 and Top-5 accuracies than single precision floats. On the other hand, when dealing with a more complex dataset as CIFAR-10, there is a loss of $12\%$ in Top-1 and $4\%$ in Top-5. In order to mitigate this, a hybrid posit-float architecture has been considered. Following the idea described in \cite{Johnson2018}, only the additions are performed in posit format. Under this scenario the accuracies are higher, but still floating point is superior. Finally, it must be emphasized that the posits employed in these tests are \positenv{8}{0}, which in exchange for the accuracy loss can reduce the memory footprint to a quarter in comparison to single precision, the functional units complexity and so on.

\section{Conclusions}
\label{sec:conclusions}

In this work an algorithm for performing multiplication between two posit numbers has been presented. The algorithm is generic for whatever \positenv{n}{es} configuration and it has been integrated into the FloPoCo framework. Furthermore, this multiplication algorithm, together with other posit operations, has been employed in the neural networks scenario for performing training and inference, obtaining promising trade-off results.

In the future, further studies must be made in order to integrate posits to train larger NNs, such as CNNs, and to perform inference with higher accuracies. A possible direction may be combining different formats, as \cite{Johnson2018} proposed, bracing the transprecision concept. 

\bibliographystyle{unsrt}  
\bibliography{references}  






\end{document}